\pdfoutput=1

\documentclass[11pt]{article}

\usepackage{acl}

\usepackage{times}
\usepackage{latexsym}
\usepackage{booktabs}
\usepackage{colortbl}
\usepackage{pythonhighlight}
\usepackage{xspace}
\usepackage{graphicx}
\usepackage{pifont}
\definecolor{lp}{HTML}{CBC3E3}

\usepackage[T1]{fontenc}

\usepackage[utf8]{inputenc}

\usepackage{microtype}

\usepackage{inconsolata}

\usepackage{graphicx}
\usepackage{kotex}

\usepackage{multirow}
\usepackage{listings}
\usepackage{amssymb}
\usepackage{amsmath}

\newcommand\blfootnote[1]{%
  \begingroup
  \renewcommand\thefootnote{}\footnote{#1}%
  \addtocounter{footnote}{-1}%
  \endgroup
}

\title{LP Data Pipeline: Lightweight, Purpose-driven Data Pipeline \\for Large Language Models}

\author{Yungi Kim$^{*}$, Hyunsoo Ha$^{*}$, Seonghoon Yang$^{*}$, Sukyung Lee, Jihoo Kim, Chanjun Park$^{ \dagger}$ \\
\\
  Upstage AI \\
  \texttt{\{eddie, hyunsooha, hoonyang, sukyung, jerry, chanjun.park\}@upstage.ai}}

\begin{document}
\maketitle
\begin{abstract}
\blfootnote{$^*$Equal Contribution $^\dagger$ Corresponding Author}
Creating high-quality, large-scale datasets for large language models (LLMs) often relies on resource-intensive, GPU-accelerated models for quality filtering, making the process time-consuming and costly. This dependence on GPUs limits accessibility for organizations lacking significant computational infrastructure. To address this issue, we introduce the \emph{Lightweight, Purpose-driven (LP) Data Pipeline}, a framework that operates entirely on CPUs to streamline the processes of dataset extraction, filtering, and curation. Based on our four core principles, the LP Data Pipeline significantly reduces preparation time and cost while maintaining high data quality. Importantly, our pipeline enables the creation of purpose-driven datasets tailored to specific domains and languages, enhancing the applicability of LLMs in specialized contexts. We anticipate that our pipeline will lower the barriers to LLM development, enabling a wide range of organizations to access LLMs more easily.
\end{abstract}

\section{Introduction}
The rapid advancement of large language models (LLMs) has significantly increased the demand for massive, high-quality datasets for effective training~\cite{zhao2023survey,liu2024datasets}. While traditional data processing and filtering methods have relied on CPU-based techniques~\cite{wenzek2019ccnet, penedo2023refinedweb, gunasekar2023textbooks, li2023textbooks}, there is a growing trend toward utilizing GPU-accelerated models to extract \textit{higher-quality data} suitable for advanced LLM training~\cite{penedo2024fineweb,li2024scalingfilter,sachdeva2024train}.

A prominent example employing GPU-based models for quality filtering is FineWeb-edu~\cite{penedo2024fineweb}. While effective for enhancing LLM performance, this approach renders the data curation process both time-consuming and costly. Consequently, organizations with limited computing resources face significant barriers to developing LLMs, restricting such advancements to well-funded entities. This scenario highlights an urgent need for effective and scalable solutions capable of extracting \textit{high-quality data} without heavy reliance on GPUs.

Meanwhile, in real-world applications, there is an escalating demand for purpose-driven LLMs tailored to specific domains and languages~\cite{ling2023domain,zheng2023lmsys,zhao2024large,qin2024multilingual}. To this end, many organizations require specialized datasets for particular industries, such as finance~\cite{lee2024survey}, law~\cite{lai2024large}, and healthcare~\cite{he2023survey}, or for underrepresented languages like Thai~\cite{kim2024representing}, Vietnamese~\cite{tran2024lavy}, and Arabic~\cite{huang2023acegpt}.

To meet these needs, we introduce the Lightweight, Purpose-driven (LP) Data Pipeline. This pipeline is a fully CPU-based framework designed to streamline the extraction, filtering, and curation of large-scale datasets tailored for specific domains and languages. By developing and employing a lightweight yet effective model for quality filtering, it eliminates the dependency on GPU-accelerated models \textit{without compromising data quality}. Furthermore, by leveraging FastText~\cite{bojanowski2017enriching} for domain classification, LP Data Pipeline facilitates the creation of specialized datasets that meet the unique requirements of different LLM applications.

We conduct empirical studies on the LP Data Pipeline to demonstrate the time and cost required to process a single CommonCrawl dump, and further conduct scalability experiments on large-scale data dumps to showcase the pipeline's capability.
Furthermore, by automating curation processes through workflow management tools like Airflow\footnote{\url{https://airflow.apache.org/}}, the LP Data Pipeline ensures that datasets remain up to date. Our pipeline effectively lowers the barrier to entry for LLM development, providing a practical solution for organizations aiming to build specialized language models efficiently and effectively.

\section{Related Work}
\subsection{Web Data Curation for LLMs}
The rapid advancement of LLMs has led to an increased demand for extensive, high-quality datasets~\cite{wenzek2019ccnet, penedo2023refinedweb, gao2020pile}. Prominent datasets such as The Pile~\cite{gao2020pile}, C4~\cite{2020t5}, RedPajama~\cite{together2023redpajama}, and SlimPajama~\cite{cerebras2023slimpajama} have been instrumental in providing vast amounts of textual data from sources like CommonCrawl~\cite{ccdump}, Wikipedia~\cite{wikidump}, and academic publications~\cite{clement2019use}. However, these datasets have the limitation of containing a significant amount of low-quality data~\cite{caswell2021quality, dodge2021documenting}. To address this limitation, Fineweb-edu~\cite{penedo2024fineweb} utilized extensive GPU resources for quality filtering, demonstrating that training LLMs with this dataset improves their performance compared to the previously mentioned datasets. While effective, this GPU-based method is time-consuming and costly due to the high computational demands involved in processing \textit{massive web corpora}~\cite{thompson2020computational}. This reliance on GPU resources poses challenges for organizations with limited computational resources, restricting their ability to develop LLMs tailored to specific domains or languages.

\subsection{Quality Filtering Techniques}
Data quality is critical for effective LLM training, as it directly impacts model performance and reliability~\cite{penedo2024fineweb,du2024data,rejeleene2024towards}.
To remove low-quality content, rule-based methods are utilized, relying on metadata attributes such as text length, stop word fraction, and n-gram repetition~\cite{together2023redpajama}. Furthermore, KenLM~\cite{heafield2011kenlm} and FastText~\cite{joulin2016bag} models are employed to filter out low-content data that remains even after rule-based methods have been applied, all \textit{without imposing heavy computational requirements}. However, we note that the studies employing these models for quality filtering simply used models already trained on Wikipedia data, \textit{without making any effort to enhance their performance}~\cite{meghwal2020can}. Recently, there has been significant research on using GPU-based models to curate higher-quality data~\cite{penedo2024fineweb,sachdeva2024train}, which leads to increased processing times and costs.
Meanwhile, in order to eliminate redundancy at the document level, deduplication techniques such as MinHashLSH~\cite{indyk1998approximate} are applied. However, these methods may not adequately address intra-document redundancy~\cite{khan2024lshbloom}.

\begin{figure*}
    \centering
    \includegraphics[width=1\linewidth]{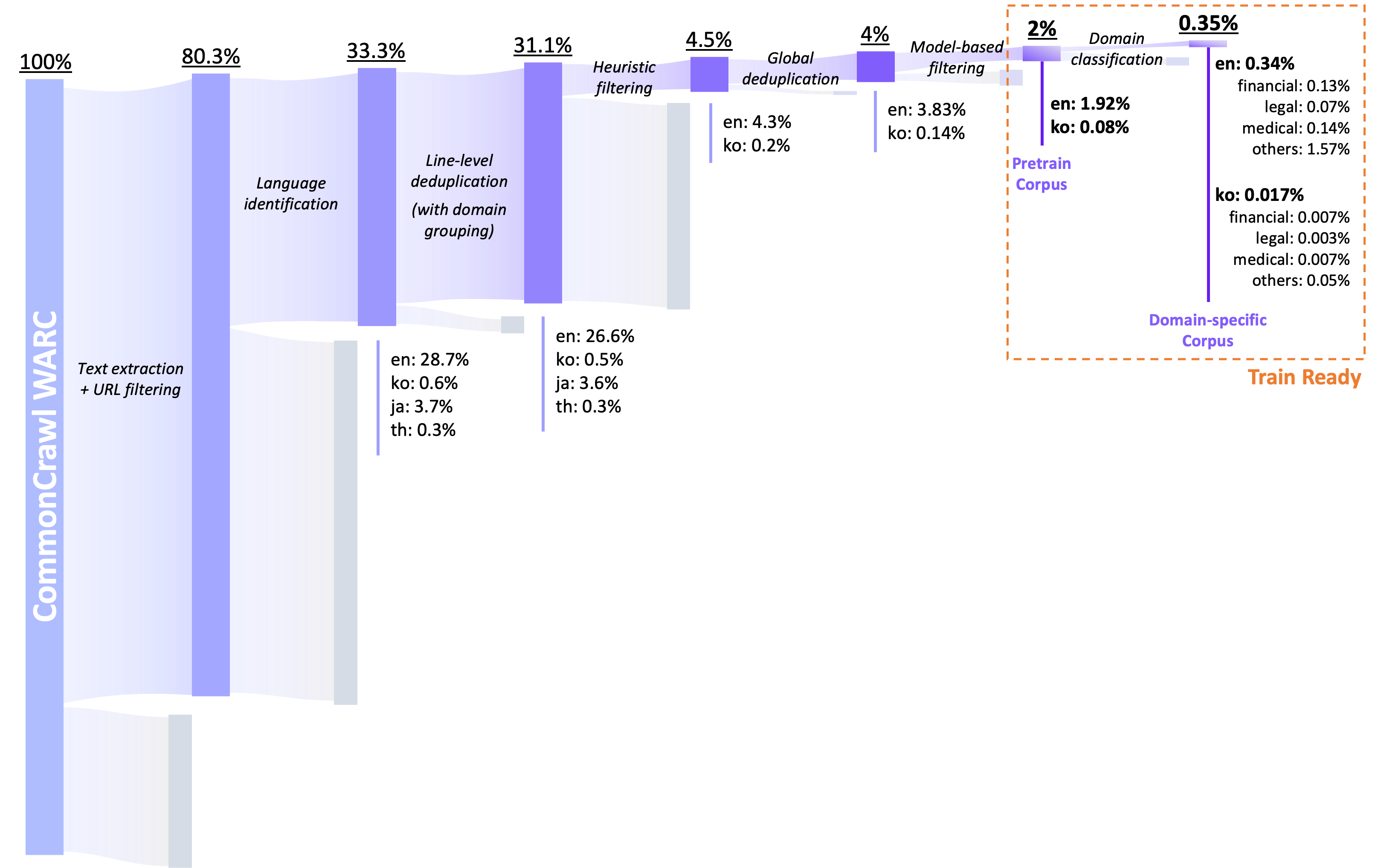}
    \caption{Overview of the \emph{Lightweight, Purpose-driven (LP) Data Pipeline}: Data extraction and cleansing flow from Common Crawl WARC dumps, illustrating the filtering processes and the sizes of data being filtered at each stage.}
    \label{fig:overview}
\end{figure*}

\section{Lightweight, Purpose-driven (LP) Data Pipeline}
\subsection{Pipeline Philosophy}
The LP Data Pipeline is founded on four core principles designed to address the challenges of curating high-quality, large-scale datasets.

\paragraph{Fully CPU-Based Data Curation.}
Recent quality filtering often depends on GPU-intensive models, which demand significant computational resources. The LP Data Pipeline mitigates this issue by utilizing optimized and efficient CPU-based methods, achieving practical performance without the need for expensive GPU infrastructure.

\paragraph{Optimized Processing Order for Efficiency.}
To enhance resource efficiency, LP Data Pipeline employs a strategic sequence for computational tasks. Initial filtering and basic data cleansing are conducted first, reserving more resource-intensive operations for later stages. This order allows only \textit{relatively} high-quality data to undergo advanced processing, thus reducing overall resource expenditure and time.

\paragraph{Continuous Knowledge Updating.}
The LP Data Pipeline includes automated mechanisms for continuous knowledge updating. By detecting and processing regular data dumps from sources such as Wikipedia and CommonCrawl, the pipeline updates its data in a timely manner without manual intervention. Workflow management tools like Airflow automate job scheduling and repetitive tasks, ensuring that LLMs trained on the pipeline's results align with the latest available information.

\paragraph{Purpose-Driven Datset Construction.}
The pipeline supports the creation of datasets tailored to specific domains, such as finance, law, and medicine. With advanced module for domain classification, it extracts data that aligns with specialized applications. For example, datasets for legal analysis can be enriched with case law references and terminology. This tailored approach enhances the performance of LLMs in real-world applications, providing domain-specific expertise and nuanced understanding.

\subsection{Curation Pipeline} 
The LP Data Pipeline is organized as a sequential process optimized for efficient computation. Each stage incrementally refines the dataset, ensuring that the resulting dataset is both high-quality and aligned with specific language and domain requirements. The following subsections provide a detailed explanation of each phase. The overview of LP Data Pipeline is illustrated in Figure~\ref{fig:overview}.

\subsubsection{Raw Text Extraction and URL Filtering}
The pipeline begins with the extraction of raw text from large-scale web data sources such as CommonCrawl. Utilizing efficient HTML parsing tools like \texttt{Resiliparse}~\cite{bevendorff:2018}, LP Data Pipeline extracts textual content while removing boilerplate elements and irrelevant HTML tags. URL filtering is applied to exclude undesirable sources based on a predefined list of domains containing harmful content~\cite{christou2020phishing}.

\subsubsection{Language Identification}
To generate language-specific data, the LP Data Pipeline employs a CPU-optimized language identification tool based on FastText~\cite{joulin2016bag}. This allows for accurate and efficient detection of the target languages (\textit{e.g.}, English, Korean, Japanese, Thai), ensuring that only the relevant textual data is retained for further processing.

\subsubsection{Line-Level Deduplication with Domain Grouping}
Redundant boilerplate text and recurring promotional phrases within documents can introduce noise and diminish the quality of training data for LLMs. To this end, common approach involves partitioning all text lines from the entire documents into \textit{random} buckets and performing comprehensive line-level deduplication across all text lines within each bucket~\cite{dubey2024llama}. In this approach, using fewer buckets leads to higher filtering performance but incurs greater computational costs for processing each bucket~\cite{shen2023slimpajama}. Conversely, increasing the number of buckets reduces the computational costs for processing each one, but inevitably lowers the filtering performance.

To address this issue, we introduce a more efficient solution through line-level deduplication \textit{using domain grouping instead of random grouping}. This method groups documents by their domain URLs and performs line-level deduplication within each domain group. This approach effectively removes boilerplate content that text extraction tools like Resiliparse may overlook and eliminates repetitive advertising phrases embedded in documents from specific sites. This line-level deduplication with domain grouping significantly reduces computational costs, offering a practical alternative to exhaustive line-level deduplication.

\begin{table}[t]
  \centering
  \resizebox{0.48\textwidth}{!}{
      \begin{tabular}{l|l}
        \toprule
        \textbf{Metadata Description} & \textbf{Threshold} \\
        \midrule
        Number of lines in the document & $\geq 5$ \\
        Length of document (in characters) & $\geq 200$ \\
        Fraction of lines starting with a bullet point & $\leq 0.9$ \\
        Fraction of lines ending with terminal punctuation & $> 0.12$ \\
        Number of sentences in the document & $\geq 5$ \\
        Fraction of words with no alphabetic characters & $\leq 0.2$ \\
        Fraction of words containing newline characters & $\leq 0.3$ \\
        Fraction of short lines in the document & $< 0.67$ \\
        Fraction of stop words in the document & $\geq 0.0$ \\
        Fraction of top 2-gram characters & $\leq 0.2$ \\
        Fraction of top 3-gram characters & $\leq 0.18$ \\
        Fraction of top 4-gram characters & $\leq 0.16$ \\
        Fraction of duplicate 5-gram characters & $\leq 0.15$ \\
        Fraction of duplicate 6-gram characters & $\leq 0.14$ \\
        Fraction of duplicate 7-gram characters & $\leq 0.13$ \\
        Fraction of duplicate 8-gram characters & $\leq 0.12$ \\
        Fraction of duplicate 9-gram characters & $\leq 0.11$ \\
        Fraction of duplicate 10-gram characters & $\leq 0.1$ \\
        Number of words in ldnoobw list & $= 0$ \\
        Word count in the document & $\geq 50$ and $\leq 100,000$ \\
        Symbol-to-word ratio in the document & $\leq 0.1$ \\
        Presence of "lorem ipsum" text & $= 0$ \\
        Fraction of lines ending with ellipsis & $\leq 0.3$ \\
        Presence of curly brackets & $= 0$ \\
        Average word length in the document & $\geq 3$ and $\leq 10$ \\
        Presence of license string & $= \text{False}$ \\
        Presence of personally identifiable information & $= \text{False}$ \\
        \bottomrule
      \end{tabular}
  }
  \caption{
    Metadata thresholds and descriptions for quality filtering in English documents.
  }
  \label{tab:thresholds}
\end{table}

\subsubsection{Heuristic Filtering} 
The LP Data Pipeline generates comprehensive metadata for each document, including metrics such as text length, stop word ratio, and n-gram repetition. The selection of metadata features was guided by established methodologies from prior studies, including C4~\cite{2020t5}, Gopher~\cite{rae2021scaling}, the Pretrainer's Guide~\cite{longpre2023pretrainer}, RefinedWeb~\cite{penedo2023refinedweb}, and Fineweb~\cite{penedo2024fineweb}. Furthermore, to ensure uphold privacy standards, the LP Data Pipeline generates metadata for Personally Identifiable Information (PII).

Based on this metadata, rule-based filtering is then applied to exclude data that do not meet predefined quality thresholds. The thresholds were determined through qualitative assessments to ensure the exclusion of substandard data. Given the extensive size and accessibility of web corpora, stringent thresholds were implemented to ensure a high standard of data quality. Table~\ref{tab:thresholds} presents the metadata and thresholds used for English documents in the LP Data Pipeline.

\subsubsection{Global Deduplication}
To mitigate redundancy within the dataset, the LP Data Pipeline utilizes MinHash Locality-Sensitive Hashing (LSH)~\cite{indyk1998approximate} to achieve efficient document-level deduplication. This technique is particularly suitable for deployment in cluster environments such as Apache Spark~\cite{spark}, enabling the effective identification and removal of near-duplicate documents. This global deduplication was conducted independently for each CommonCrawl dump.

\begin{figure*}
    \centering
    \includegraphics[width=0.8\linewidth]{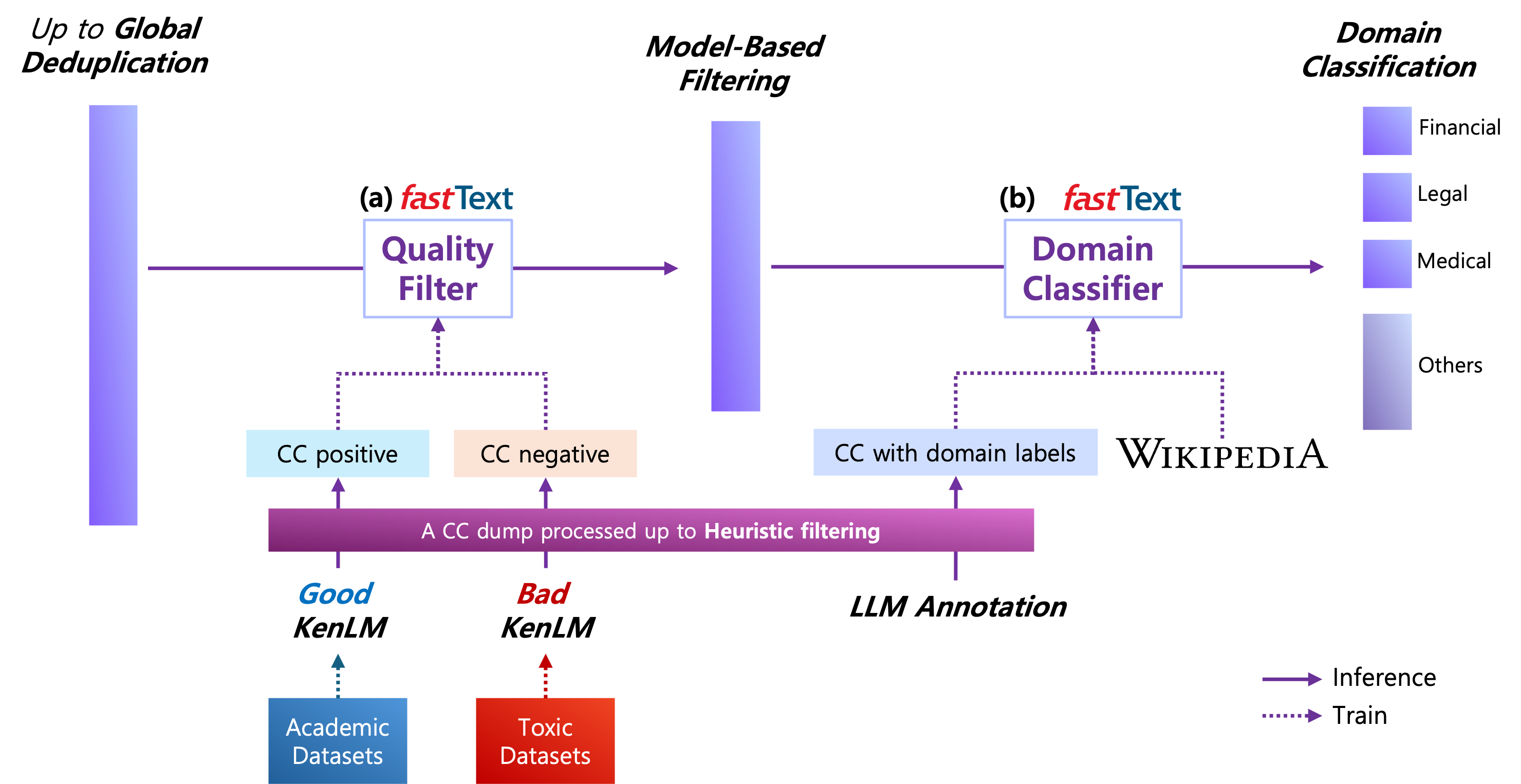}
    \caption{Overview of the training process for the quality filtering model and the domain classification model.}
    \label{fig:training_overview}
\end{figure*}

\subsubsection{Model-based Quality Filtering}
Measuring data quality is essential in curating datasets for training LLMs. While GPU-based reward models can achieve strong performance, they suffer from low throughput and incur substantial costs when processing massive datasets like Common Crawl, making them impractical. To facilitate the curation of high-quality data using only CPU computation, we enhance the performance of the FastText model by utilizing datasets generated through both Good KenLM and Bad KenLM~\cite{kim2024rethinking}. Figure~\ref{fig:training_overview}-(a) illustrates our training overview.

The traditional KenLM (\textit{i.e.}, Good KenLM) is trained on high-quality sources such as Wikipedia and academic textbooks to identify data with reliable and informative content. This helps distinguish documents that contribute positively to LLM training. On the other hand, Bad KenLM is trained on low-quality content, including toxic language, hate speech, and less educational materials like informal social media posts. This training allows Bad KenLM to identify undesirable data that could negatively impact model performance. A FastText classifier is trained on CommonCrawl samples labeled as positive by Good KenLM and negative by Bad KenLM, and is then utilized.

This strategy utilizes Good and Bad KenLMs to construct FastText training data, in contrast to existing methodologies that typically use Wikipedia articles as positive samples and random documents as negative samples. This innovative strategy enables more effective data quality assessment using CPU-based resources, paving the way for effective and scalable LLM data curation.

\subsubsection{Domain Classification}
Following the work of \citet{cheng2024adaptinglargelanguagemodels}, training LLMs with domain-specific data has proven to substantially enhance their performance within targeted domains. To curate such data, we developed a FastText~\cite{joulin2016bag} model trained to classify documents into three industrial domains: finance, law, and healthcare.\footnote{Additional domains can be incorporated with ease.} Figure~\ref{fig:training_overview}-(b) illustrates our training overview.

To construct the training data, we employed the steps previously outlined in the LP Data Pipeline, including model-based filtering, applied to a CommonCrawl dump~\cite{ccdump}. The resulting data was annotated using a LLM, categorizing documents into the three primary industrial domains with an additional `others' label for documents that did not align with these categories, culminating in four classes in total. In addition, some documents were also extracted from the Wikipedia Cirrus dump~\cite{wikidump} by specifically identifying categories with keywords closely related to each industrial domain.

The FastText model, trained on this dataset, facilitates the curation of domain-specialized datasets that enhance LLM performance in targeted domains.

\begin{table}[t]
  \centering
  \resizebox{0.48\textwidth}{!}{
      \begin{tabular}{c|cc}
        \toprule
        \textbf{Processing Phase} & \textbf{Processing Time} & \textbf{Estimated Cost} \\
        \midrule
        Raw Text Extraction & 1h 30m & \$128.69 \\
        Language Identification & 19m & \$20.7 \\
        Line-Level Deduplication & 1h & \$85.79 \\
        Heuristic Filtering & 25m & \$27.24 \\
        Global Deduplication & 13m & \$14.16 \\
        Model-based Quality Filtering & 48m & \$68.63 \\
        Domain Classification & 7m & \$7.62 \\
        \midrule
        \textbf{Total} & \textbf{4h 22m} & \textbf{\$352.83} \\
        \bottomrule
    \end{tabular}
  }
  \caption{Processing time and estimated cost for each phase of the CC-MAIN-2024-10 dump file analysis, showcasing the efficiency of the LP Data Pipeline in handling large-scale data processing.}
  \label{tab:processing}
\end{table}

\section{Empirical Analysis}
\subsection{Practical Feasibility of the LP Data Pipeline}
To validate the feasibility of the LP Data Pipeline with large-scale datasets, we conducted experiments using actual CommonCrawl dumps, focusing on key metrics such as processing time and estimated cost. The results demonstrate the pipeline’s efficiency and cost-effectiveness in handling large-scale datasets. Table~\ref{tab:processing} provides a breakdown of each processing phase for the 4TB CC-MAIN-2024-10 CommonCrawl dataset, processed on 128 machines with 8-core CPUs. The entire process was completed in just 4 hours and 22 minutes, at an estimated cost of \$352.83\footnote{The estimated cost was calculated using 120 AWS m7a.2xlarge~\cite{AWSM7aInstances} instances as the basis.}, highlighting the pipeline’s suitability for large-scale data curation.

The raw text extraction phase was the most intensive, providing the foundation for subsequent tasks. Language identification filtered relevant data, followed by line-level deduplication to enhance data quality by removing redundant content. Heuristic filtering and global deduplication further refined the dataset, while domain model-based quality filtering and classification ensured high data standards and facilitated targeted content selection, respectively. This result affirms that the LP Data Pipeline is a practical and cost-efficient solution for constructing large-scale, high-quality datasets using CPU-based resources, making it accessible to organizations with limited GPU infrastructure.

\subsection{Quantity of Domain-Specific Datasets}
Table~\ref{tab:token} presents the results from processing the CC-MAIN-2024-10 dump into domain-specific datasets for English and Korean. The Medical Corpus included 5.02 million English documents and 0.23 million Korean documents, amounting to 4.27 billion and 0.24 billion tokens, respectively. Similarly, the Financial Corpus comprised 4.29 million English and 0.19 million Korean documents, with corresponding token counts of 3.99 billion and 0.20 billion, respectively. The Legal Corpus showed a smaller scale, with 2.18 million English and 0.05 million Korean documents, totaling 2.02 billion and 0.06 billion tokens, respectively. The Common Corpus, which does not correspond to the three corpora mentioned above, contained the highest volume, with 52.11 million English and 1.34 million Korean documents, contributing 46.35 billion and 1.73 billion tokens, respectively. These results are consistent with the expectation that legal documents are the least common type of content on the web.

\begin{table}[t]
  \centering
  \resizebox{0.48\textwidth}{!}{
      \begin{tabular}{c|cc|cc}
        \toprule
        \textbf{} & \multicolumn{2}{c|}{\textbf{Number of Documents}} & \multicolumn{2}{c}{\textbf{Token Count}} \\
        \textbf{Dataset} & \textbf{English} & \textbf{Korean} & \textbf{English} & \textbf{Korean} \\
        \midrule
        Medical Corpus & 5.02M & 0.23M & 4.27B & 0.24B \\
        Financial Corpus & 4.29M & 0.19M & 3.99B & 0.20B \\
        Legal Corpus & 2.18M & 0.05M & 2.02B & 0.06B \\
        Common Corpus & 52.11M & 1.34M & 46.35B & 1.73B \\
        \bottomrule
    \end{tabular}
  }
  \caption{Quantity of domain-specific datasets from the CC-MAIN-2024-10 dump, showing document counts and token volumes for English and Korean, respectively.}
  \label{tab:token}
\end{table}

\begin{figure}[!t]
    \centering
    \includegraphics[width=1.0\linewidth]{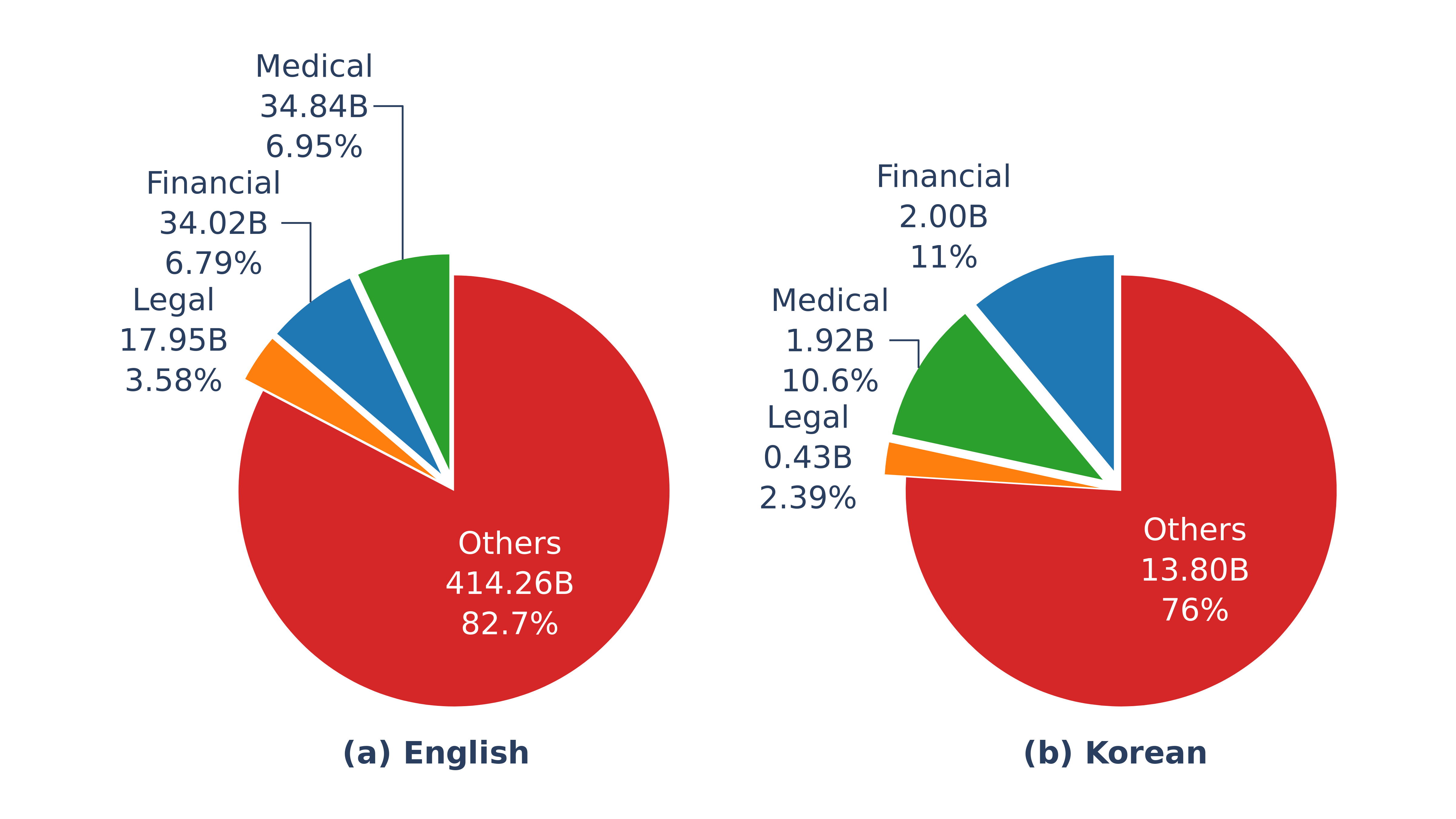}
    \caption{
        The token distribution of domain-specific datasets obtained from processing 10 CommonCrawl dumps for English and Korean using the \emph{Lightweight, Purpose-driven (LP) Data Pipeline}.
    }
    \label{fig:piechart}
\end{figure}

\subsection{Scaling Dump Analysis}
We expanded our analysis to include 10 CommonCrawl dumps, processed using the LP Data Pipeline. Figure~\ref{fig:piechart} illustrates the token distribution of domain-specific datasets for English and Korean, respectively. For the English datasets, the analysis yielded 17.95 billion tokens for the Legal domain, 34.02 billion for Finance domain, and 34.84 billion for Medical domain. For the Korean datasets, the results showed 2 billion tokens for the Finance domain, 1.92 billion for Medical domain, and 0.43 billion for Legal domain. These results are similar to the trends observed in a single dump, where legal documents are the least prevalent type of content.
Furthermore, these results emphasize that the LP Data Pipeline can effectively build extensive, domain-specific datasets for both major and low-resource languages like Korean, offering an advantage over traditional web crawling methods.
The total time and cost for processing 10 dumps can be approximated by scaling the results of a single dump by a factor of 10, highlighting our pipeline’s scalability and cost-efficiency for comprehensive data curation.

\section{Conclusion}
In this paper, we introduce the \textit{Lightweight, Purpose-driven (LP) Data Pipeline}, a fully CPU-based framework designed for efficient curation of high-quality datasets for LLM training. Based on our four core principles, the LP Data Pipeline significantly reduces computational costs while maintaining data quality comparable to that of GPU-based methods. Our empirical analysis demonstrated that the LP Data Pipeline is highly cost-effective and time-efficient, and scales efficiently with large datasets. For future work, we plan to conduct LLM training experiments with the curated datasets to further validate the effectiveness of our pipeline.

\section*{Acknowledgments}
This work was supported by Institute of Information \& Communications Technology Planning \& Evaluation(IITP) grant funded by the Korea government(MSIT) (No. RS-2024-00338140, Development of learning and utilization technology to reflect sustainability of generative language models and up-to-dateness over time).

\section*{Limitations}
The \textit{Lightweight, Purpose-driven (LP) Data Pipeline} offers an efficient CPU-based solution for dataset curation but has certain limitations. A primary limitation is that the current language and domain classification is restricted to a subset of languages (English, Korean, Japanese, and Thai) and specific domains (finance, law, and healthcare). Expanding the framework to include additional languages and domains, especially low-resource ones, would greatly improve its versatility and applicability for various LLM training needs.

Additionally, due to resource constraints, extensive LLM training experiments using the curated datasets were not conducted. This limits our ability to empirically validate the impact of the LP Data Pipeline on model performance. Future work should involve comprehensive evaluations to thoroughly assess how the curated datasets influence LLM training outcomes.

\section*{Ethics Statement}
The \textit{Lightweight, Purpose-driven (LP) Data Pipeline} was developed with a commitment to ethical standards, emphasizing data privacy, legality, and fairness. The pipeline incorporates measures for detecting and removing PII to safeguard individual privacy and ensures adherence to copyright and licensing requirements.

\bibliography{custom}

\clearpage
\appendix
\onecolumn

\section{System Architecture}
\begin{figure*}[!t]
    \centering
    \includegraphics[width=1.0\linewidth]{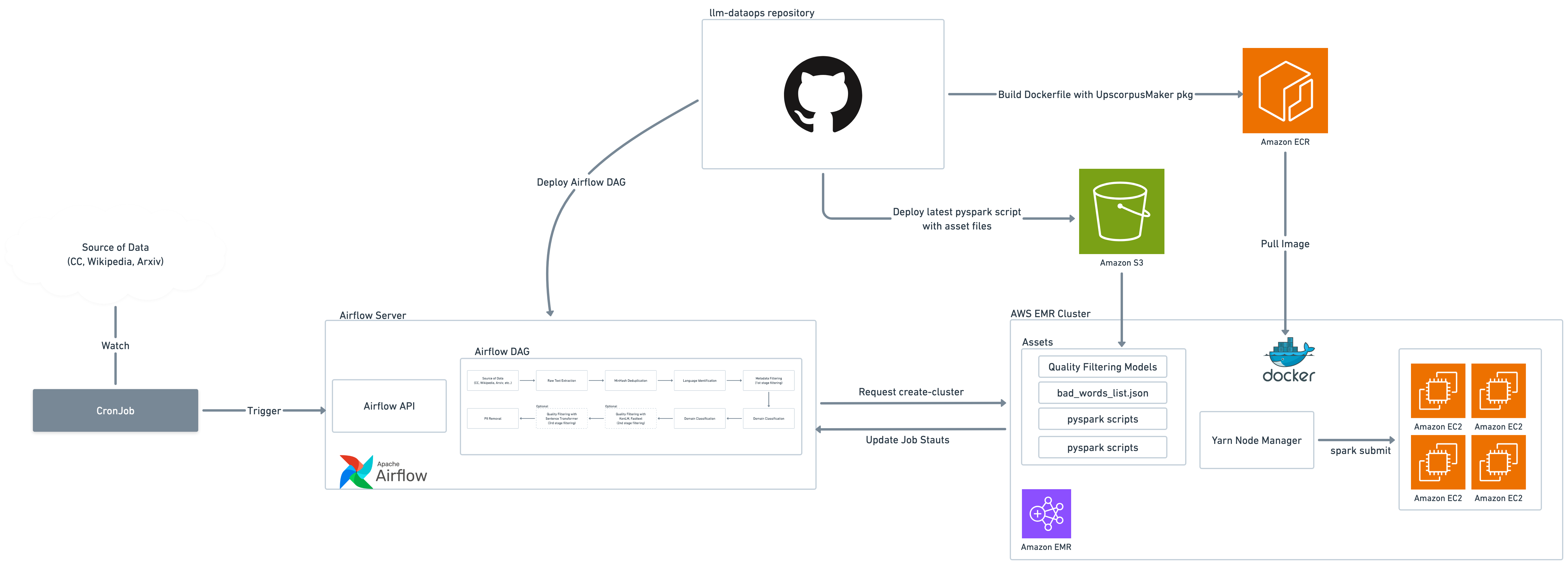}
    \caption{System Architecture of the \emph{Lightweight, Purpose-driven (LP) Data Pipeline}}
    \label{fig:architecture}
\end{figure*}

Figure~\ref{fig:architecture} presents the system architecture of the Lightweight, Purpose-driven (LP) Data Pipeline. The overall management of the pipeline is orchestrated by Airflow, which employs a cron job to periodically check for new data releases from sources such as Wikipedia, CommonCrawl, and ArXiv. Upon detecting new data releases, Airflow initiates the execution of predefined Directed Acyclic Graphs (DAGs) that perform tasks including text extraction, deduplication, filtering, and domain classification. These processes are executed on a Spark cluster deployed within AWS EMR, which is configured to leverage Docker images built via GitHub CI/CD pipelines and utilizes quality filtering and domain classification models stored in S3. This architectural design supports the continuous collection and processing of up-to-date data with minimal human intervention, thereby automating the construction of train-ready, purpose-driven datasets.

\section{Results of Line-Level Deduplication with Domain Grouping}

\begin{table}[h]
  \centering
  \resizebox{0.55\linewidth}{!}{
      \begin{tabular}{l|l}
        \toprule
        \textbf{Removed Lines (English)} & \textbf{Removed Lines (Korean)} \\
        \midrule
        Leave a Comment & 본문 바로가기 \\
        Skip to Main Content & 콘텐츠로 건너뛰기 \\
        Last Updated & 답글 남기기 \\
        Scroll to Top & 이 기사를 공유합니다 \\
        Share & 자주 묻는 질문 \\
        \bottomrule
    \end{tabular}
  }
  \caption{Top five lines removed from English and Korean documents using the line-level deduplication method with domain grouping.}
  \label{tab:linededup}
\end{table}

Table~\ref{tab:linededup} presents  the top five lines removed from both English and Korean documents through line-level deduplication. While domain grouping significantly reduces computational overhead, our findings confirm that the method effectively removes boilerplate content as intended.

\newpage

\section{Data Samples}
This section presents representative data samples in both Korean and English, sourced from the CC-MAIN-2024-38 dump, covering the finance, law, and medical domains. Table \ref{tab:legal_sample_en} presents an English legal text sample, while Table \ref{tab:legal_sample_ko} provides a Korean legal text sample. Similarly, Table \ref{tab:medical_sample_en} presents an English medical text sample, and Table \ref{tab:medical_sample_ko} includes a Korean medical text sample. For the finance domain, Table \ref{tab:finance_sample_en} presents an English sample, with Table \ref{tab:finance_sample_ko} providing a Korean sample.

\begin{table}[h!]
\centering
\begin{tabular}{|p{0.9\textwidth}|}
\hline
\textbf{Legal English Sample} \\ \hline
A Guide to International Law

International law is the set of rules accepted by various countries, binding them in agreement to what policies they will all follow. International rules between countries will serve as a framework for the practice of international relations.

Unlike most other areas of law, international law has no defined area of governing body. The piecemeal collection of international law will encompass customs, agreements, treaties, tribunals, legal precedents from international courts and more.

International law varies from state-based legal systems in that it is not applied on an individual basis, but to countries as a whole. The first step to creating an international law is to obtain jurisdiction over a country. The Geneva Convention is the main source of international laws, which are enforced by international courts. These laws will be treated like your traditional national law when a country signs a specific treaty. The difficult aspect about this is that a country will need to consent to be governed by a law.

Although there is no definitive governing body overseeing international law, the United Nations is the most widely recognized and influential international organization, with the International Court of Justice being its judicial counterpart.

Public international laws are concerned with questions of rights between nations or nations and the citizens of other nations. On the other hand, private international law deals with controversies between private persons or natural born citizens that have significant relationships to more than one nation. Additionally, there are always the international business laws between companies that do business in more than one country.

International law will cover basic concepts that every national legal system will such as property, torts, procedure and remedies. However, the main substantive law is: international economic law, security law, criminal law, environmental law, diplomatic law, humanitarian and human rights law.

International laws cover a myriad of legal concepts coming from a number of sources. Understanding what customs or national laws as well as what international laws will apply to your business or even as an individual traveling is essential.\\ 
\hline
\end{tabular}
\caption{Example of an English legal text}
\label{tab:legal_sample_en}
\end{table}

\begin{table}[h!]
\centering
\begin{tabular}{|p{0.9\textwidth}|}
\hline
\textbf{Legal Korean Sample} \\ \hline
기소유예정보 알아보기

기소유예: 단순하게 설명한 법률 용어

법률 용어는 종종 복잡하고 이해하기 어렵습니다. 그 중 하나인 ‘기소유예’에 대해 쉽게 알아봅시다.

기소유예란 무엇인가?

기소유예는 검찰이 피의자를 기소할 수 있는 권리를 일정 기간 동안 유예하는 것을 말합니다. 이는 피의자가 처음 범죄를 저지른 경우나 경미한 범죄에 대해 적용되며, 이 기간 동안 피의자가 다시 범죄를 저지르지 않는다면 기소권이 소멸되어 처벌받지 않게 됩니다.

기소유예의 장점

기소유예는 피의자에게 두 번째 기회를 줄 수 있는 좋은 방법입니다. 피의자가 범죄를 저지르지 않고 사회에 잘 적응하면, 그들에게 불이익이 가해지지 않을 수 있습니다. 이는 범죄 재발률을 줄이는 데에도 도움이 됩니다.

기소유예의 단점

그러나 기소유예는 범죄를 저지른 사람이 처벌을 피하게 만들 수도 있습니다. 이는 피해자나 사회에 대한 부정적인 메시지를 보낼 수 있습니다. 따라서 검찰은 기소유예를 결정할 때 신중해야 합니다.\\ 
\hline
\end{tabular}
\caption{Example of a Korean legal text}
\label{tab:legal_sample_ko}
\end{table}

\begin{table}[h!]
\centering
\begin{tabular}{|p{0.9\textwidth}|}
\hline
\textbf{Medical English Sample} \\ \hline
Otosclerosis

Otosclerosis is an abnormal growth of bone in the middle ear that causes hearing loss. It typically begins in the early 20s, and is the leading cause of middle ear hearing loss in young adults.

The exact cause of otosclerosis is not known, but evidence suggests a genetic link passed down from parent to child. Middle-aged Caucasian women are most at risk, and pregnancy seems to be a contributing factor, perhaps due to hormonal changes a woman is undergoing at the time. This bone growth usually occurs around the stapes bone in the middle ear, preventing it from moving freely, essential to proper hearing.

Gradually worsening hearing loss is the primary symptom of otosclerosis. It may begin with an inability to hear low-pitched sounds or whispers. Other symptoms may include vertigo or dizziness and tinnitus (ringing in the ears).

Treatments

The symptoms of otosclerosis are like those of other conditions, so a thorough examination by an otolaryngologist is essential in ruling out other problems and diagnosing the disease. A hearing test will usually show signs of conductive hearing loss in the lower frequency tones, a hallmark of otosclerosis.

Mild cases of otosclerosis can be corrected with a hearing aid designed to amplify sounds. Orally ingested sodium fluoride has been shown to slow the progression of the disease, and may be an option.

In more advanced cases, a surgical procedure known as a stapedectomy is often performed. In this surgery, part or all of the affected stapes bone is removed and replaced with a prosthetic device that enables the bones of the middle ear to resume movement, allowing sound waves to reach the inner ear, improving or restoring hearing.

There are inherent risks in any surgery, but left untreated, otosclerosis will only get worse. Speak to your doctor about the best treatment options for your hearing loss.\\ 
\hline
\end{tabular}
\caption{Example of an English medical text}
\label{tab:medical_sample_en}
\end{table}

\begin{table}[h!]
\centering
\begin{tabular}{|p{0.9\textwidth}|}
\hline
\textbf{Medical Korean Sample} \\ \hline
혈소판 이상은 혈소판에 이상이 생기는 질환을 말합니다. 혈소판은 상처가 생겼을 때 혈액을 멎게 해주는 역할을 합니다. 혈소판은 손상된 혈관벽에 붙거나 혈소판끼리 서로 엉겨 붙으면서 혈액 응고를 일으켜 혈액을 멎도록 해줍니다. 성인의 경우 혈액 1마이크로 리터 안에 약 15~40만 개의 혈소판이 있습니다. 혈소판은 혈구 중에서 크기가 가장 작으며, 골수에서 만들어집니다. 이러한 혈소판의 수나 기능에 이상이 생기면 지혈 작용에 영향을 주어 출혈이 생길 수 있습니다.

혈소판 이상증에는 혈소판 자체가 이상한 경우, 혈소판이 완전한 기능을 발휘하기 위해 필요한 혈장 성분에 이상이 있는 경우가 있습니다.
 
혈소판 기능 이상은 후천성 기능 이상과 선천성 기능 이상으로 나눌 수 있습니다. 후천성 혈소판 기능 이상에는 많은 약물에 의한 경우와 신장 부전, 간 부전, 다발성 골수종 등의 질환에 의한 경우가 있습니다.

혈소판 기능 이상을 야기하는 약물로는 비스테로이드성 소염제가 있습니다. 동맥 경화나 심장병을 예방한다고 알려져 오랫동안 인기를 누렸던 아스피린은 흔하지는 않지만 적은 용량의 복용에서도 다량의 출혈이 야기될 수 있습니다.
 
혈소판 수치가 증가하는 경우는 감염, 수술, 염증, 약물 등 여러 신체에 대한 자극에 의해 반응성으로 증가하는 2차성 혈소판 증가증이 있고, 이러한 원인 없이 골수 내에서 혈소판 생성이 자발적으로 증가하는 특발성 혈소판 증가증이 있습니다. 특발성 혈소판 증가증의 주된 원인은 혈소판 생성에 관여하는 신체의 신호 전달 과정에 이상이 생겨서 혈소판을 만들라는 명령을 계속 내리는 것입니다.

혈소판 이상에 의한 증상은 지혈 작용이 일어나지 않는다는 점입니다. 출혈 정도는 다양합니다. 출혈 경향이 나타나는 경우, 증세 없이 출혈이 일어나는 경우, 외상, 수술, 출산 시 출혈이 발생하여 발견되는 경우 등이 있습니다.

한편, 혈소판이 필요 이상으로 증가하는 경우도 발생합니다. 특발성으로 혈소판이 증가하는 경우에는 혈전증, 색전증과 같은 치명적인 합병증을 일으키기도 합니다.\\ 
\hline
\end{tabular}
\caption{Example of a Korean medical text}
\label{tab:medical_sample_ko}
\end{table}

\begin{table}[h!]
\centering
\begin{tabular}{|p{0.9\textwidth}|}
\hline
\textbf{Financial English Sample} \\ \hline
Revenue or Total Income to Profit conversion

One of the most useful tools for analysing the financial performance of a company is the Revenue or Interest Income to Profit waterfall conversion. This is a method that shows how a company makes money by graphically summarising the financial statements. It helps us to understand how a company makes money from its core business activities and how much it spends on various expenses.

TradingView solution makes it possible to use this method for two types of companies: corporate entities and financial companies. Corporate entities are businesses that sell goods or services to customers and generate revenue from sales. Financial companies are businesses that provide financial services such as lending, investing, or insurance and generate interest income from their assets.

For corporate entities, along the way we get indicators such as the Gross Profit, Operating Income, Net Income (and margins). This indicator helps investors easily understand how effective the business is currently, how much the company makes for each dollar of revenue or interest income, what it spends the most on, and where the most potential for business improvement are.

To calculate the Net Profit of a financial company like a bank, we can use a similar method as for corporate entities, but with some differences. Financial companies generate the Interest Income from their assets such as loans, securities, etc. and pay the Interest Expense on their liabilities such as deposits, borrowings, etc.. We start with the sum of Interest Income and Non-Interest Income (fees, commissions, trading, etc. ) then deduct the following items: Interest Expense and Loan Provisions (the amount of money that the company sets aside to cover potential losses from bad loans or defaults), Non-Interest Expense, Unusual Expense and Net Taxation. The outcome is the Net Income of the financial company. Along the way we can also arrive at the bank-specific intermediate stages such as the Net Interest Income After Loan Loss Provision, Bank Operating Income, and Pretax Income. This method helps us to examine the step-by-step process from the Interest Income to Net Profit and to pinpoint the main factors of bank’s non-treasury profitability.

As you can see from the Waterfall chart, each component of the income statement is shown as a bar that either adds to or subtracts from the previous bar. The final bar shows the net income or net profit of the company. This way, we can easily see how each component affects the profitability of the company and compare the margins at each stage.\\ 
\hline
\end{tabular}
\caption{Example of an English financial text}
\label{tab:finance_sample_en}
\end{table}

\begin{table}[h!]
\centering
\begin{tabular}{|p{0.9\textwidth}|}
\hline
\textbf{Financial Korean Sample} \\ \hline
자동차 취등록세에 관한 이해와 계산 방법

자동차를 구입하고 등록하려면 취등록세를 납부해야 합니다. 취등록세는 자동차의 가격에 비례하여 부과되며, 각 지역의 세율과 운행 년수에 따라 달라집니다. 이번 글에서는 자동차 취등록세에 대해 자세히 알아보고, 어떻게 계산되는지 알아보겠습니다.

취등록세의 종류

자동차 취등록세는 크게 취득세와 등록세로 나뉩니다. 취득세는 차량을 구매하는 순간에 부과되는 세금으로, 차량의 가격에 비례하여 부과됩니다. 등록세는 취득세 이후에 등록하는 과정에서 부과되는 세금으로, 차량의 운행 년수에 따라 다르게 부과됩니다.

취득세 계산 방법

취득세는 대부분의 국가에서 공식적인 공식을 통해 계산됩니다. 한국의 경우, 차량의 가격과 배기량, 연료 종류에 따라 취득세가 계산됩니다. 가격은 차량의 실제 구매 가격을 의미하며, 배기량은 차량의 엔진 용량을 나타냅니다. 연료 종류는 주로 가솔린, 디젤, LPG 등이 있으며, 연료 종류에 따라 취득세 비율이 다를 수 있습니다. 이러한 요소들을 고려하여 취득세를 계산합니다.

등록세 계산 방법

등록세는 취득세와 달리 차량의 운행 년수와 지역에 따라 달라집니다. 대부분의 국가에서는 등록세를 간단한 공식을 통해 계산하며, 운행 년수에 따라 세율이 점차 증가하거나 감소합니다. 지역에 따라서도 세율이 다르기 때문에, 동일한 차량이라도 지역에 따라 등록세가 다를 수 있습니다.

자동차 취등록세는 차량을 구입하고 등록하기 위해 부과되는 세금입니다. 취득세는 차량의 가격과 배기량, 연료 종류에 따라 계산되며, 등록세는 차량의 운행 년수와 지역에 따라 달라집니다. 이를 고려하여 자동차 취등록세를 정확하게 계산하는 것이 중요합니다. 자동차 구매 전에 취득세와 등록세를 미리 예상하여 예산을 세우고, 자세히 알아두면 자동차 구입 과정에서 도움이 될 것입니다.\\ 
\hline
\end{tabular}
\caption{Example of a Korean financial text}
\label{tab:finance_sample_ko}
\end{table}

\end{document}